%% file: main.tex
\documentclass{article} 
\usepackage{iclr2026_conference,times}

\input{math_commands.tex}

\usepackage{hyperref}
\usepackage{url}
\usepackage{svg}
\usepackage{booktabs}
\usepackage{multirow}
\usepackage{subcaption}
\usepackage{graphicx}
\usepackage{caption}

\title{Test-Time Meta-Adaptation with \\Self-Synthesis}

\author{Zeyneb N. Kaya, Nick Rui \\
Stanford University\\
\texttt{\{zeynebnk, nickrui\}@stanford.edu} 
}

\newcommand{\ours}{\textsc{MASS}}

\iclrfinalcopy 
\begin{document}

\maketitle

\begin{abstract}
As strong general reasoners, large language models (LLMs) encounter diverse domains and tasks, where the ability to adapt and self-improve at test time is valuable. We introduce \ours{}, a meta-learning framework that enables LLMs to self-adapt by generating problem-specific synthetic training data and performing targeted self-updates optimized for downstream performance at inference time. We train this behavior end-to-end via bilevel optimization: an inner loop adapts on self-generated examples while an outer loop meta-learns data-attribution signals and rewards post-update task performance. The synthetic data is optimized with scalable meta-gradients, backpropagating the downstream loss through the inner updates to reward useful generations. Experiments on mathematical reasoning show that \ours{} learns to synthesize per-instance curricula that yield effective, data-efficient test-time adaptation.

\end{abstract}

\section{Introduction}
Large language models (LLMs) have demonstrated great general-purpose capabilities, yet they are typically deployed as static artifacts. In real-world applications, however, models must continuously adapt to evolving tasks, new information, and shifting distributions encountered during deployment \citep{zheng2024towards,wu2025continual}. 

The ability of models to \textit{learn how to learn} at test time represents a critical frontier in AI development~\citep{SEAL, TTT,MAML,hu2025test}. Rather than relying on fixed pretrained capabilities, we explore whether models can learn how to best adapt themselves at test time for a new task or domain. Our approach frames this challenge as a bilevel optimization problem where models generate and curate their own problem-specific synthetic data to achieve the best self-updates at inference time. This enables models to utilize test-time compute to adapt to each unique problem they encounter and become data-efficient where high-quality task-specific supervision is sparse \citep{SelfInstruct,STaR, STP}. 

We introduce \ours{} (\textsc{\textbf{M}eta-\textbf{A}daptation with \textbf{S}elf-\textbf{S}ynthesis}), a meta-learning framework for models to self-adapt at test time. We demonstrate the effectiveness of our method in mathematical reasoning across diverse fields; by dynamically constructing a synthetic curriculum for self-improvement at test time, \ours{} addresses problem-specific knowledge gaps while offering a scalable alternative to massive offline pretraining.

\section{Methods}
\label{gen_inst}

\begin{figure}[t]
\centering
\includegraphics[scale=1]{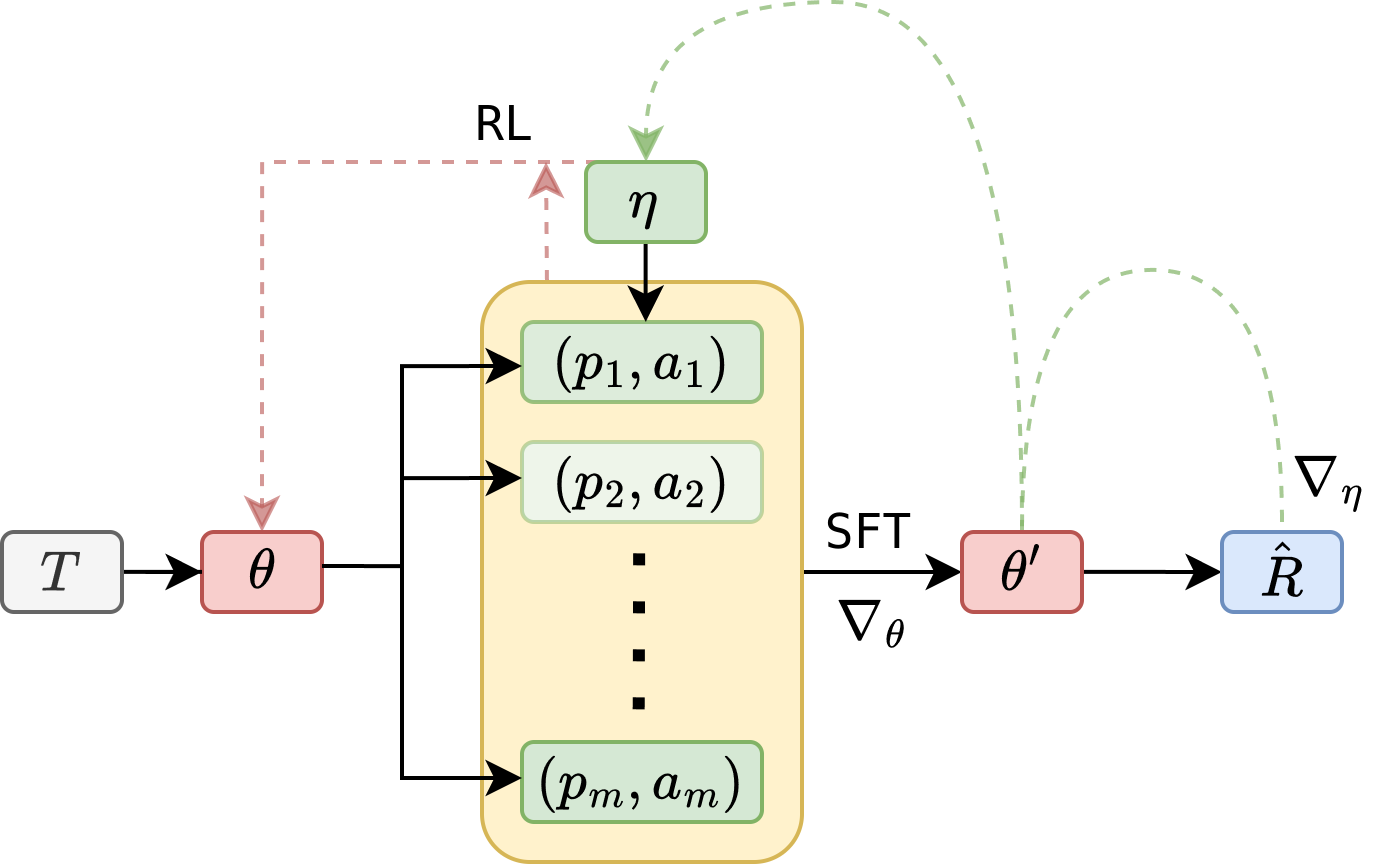}
\caption{\textbf{\ours{} pipeline.} The generator parameterized by $\theta$ generates $m$ auxiliary training examples $(p_1,a_1),\dots,(p_m,a_m)$ given the target task $T$. The examples are assigned scores using the scorer parameterized by $\eta$. Then, $\theta$ performs a weighted SFT self-update on the synthetic data to produce an adapted model $\theta'$ before producing resposne $\hat{R}$ in attempting the target problem. Meta-gradients computed by backpropagating target performance through the inner updates are used to update $\eta$ to identify examples that help the downstream loss and reward $\theta$ to generate better examples.}
\label{fig:pipeline}
\end{figure}

We formulate test-time adaptation as a bilevel meta-learning problem. Each input is treated as a distinct task, and we learn \emph{what} small synthetic self-generated dataset enables the best downstream performance for each task. Concretely, \ours{} has a \textit{generator} which generates a corpus of synthetic auxiliary problem-solution pairs given a target task and a \textit{scorer} which assigns scores to these examples based on their relevance to the target task \citep{DataRater}. During training, \ours{} then performs temporary parameter updates with this weighted dataset prior to solution generation. Optimization of the model on this downstream performance allows the scorer to learn what examples are most important and the generator to generate such examples. The overall pipeline is illustrated in Figure~\ref{fig:pipeline}.

\subsection{Data Generation and Inner-loop adaptation}
We have a generator model $\pi_\theta$ and a scorer $s_\eta$. For a target task $T$, we sample $m$ auxiliary examples $(p_i,a_i) \sim \pi_\theta(\cdot \mid T)$ for $i \in [m]$, forming an auxiliary dataset $\mathcal{D}(T) = \{(p_i,a_i)\}_{i=1}^m$. 
A scorer $s_\eta$ assigns each example a relevance weight $ s_i \;=\; s_\eta(T,p_i,a_i).$

Initializing from the current model parameters $\theta$, we perform a short update on the weighted auxiliary data according to
\[
\mathcal{L}_{\mathrm{inner}}(\theta,\eta;T)
\;=\;
\sum_{i=1}^{m} s_\eta(T,p_i,a_i)\,\ell(p_i,a_i;\theta),
\]
where $\ell(\cdot)$ is a supervised loss. The adapted model $\theta'$ is then used to attempt the target task $T$.

\subsection{Outer objective}
We optimize $\theta$ and $\eta$ so that the adapted parameters $\theta'$ achieve strong performance on $T$:
\[
\min_{\theta,\eta}\;\; \mathcal{L}_{\mathrm{outer}}\!\left(\theta'(\theta,\eta;T);\,T\right).
\]
In our experiments, we study two instantiations of $\mathcal{L}_{\mathrm{outer}}$ depending on the assumptions of the setting: if a gold solution $R^\star$ is provided for $T$ during training, we use standard cross-entropy; if no gold solution is available but we can verify candidate responses, we sample $k$ attempts $\{\hat{R}_j\}_{j=1}^k$ from $\pi_{\theta'}(\cdot\mid T)$, and treat verified responses as the target. 

\subsection{Optimization}
Backpropagating through the inner update with higher-order differentiation yields the meta-gradient for the scorer parameters $\eta$: 
\[
\frac{\partial \mathcal{L}_{\mathrm{outer}}}{\partial \eta}
=
\frac{\partial \mathcal{L}_{\mathrm{outer}}}{\partial \theta'}
\,
\frac{\partial \theta'}{\partial \eta}.
\]
Updating according to this encourages the scorer $s_\eta$ to upweight auxiliary examples whose induced adaptation step improves outer performance on $T$.
We can derive the sensitivity to each example-level score from this computation as $
\frac{\partial \mathcal{L}_{\mathrm{outer}}}{\partial s_i}
=
\left\langle
\nabla_{\theta'}\mathcal{L}_{\mathrm{outer}}(\theta';T),\;
\frac{\partial \theta'}{\partial s_i}
\right\rangle,$ which directly measures whether increasing $s_i$ would decrease the outer loss.

We use the meta-signal $-\frac{\partial \mathcal{L}_{\mathrm{outer}}}{\partial s_i}$ as a reward shaping term for the generator $\pi_\theta$ to produce auxiliary examples that are useful for adaptation: auxiliary samples that would \emph{reduce} the outer loss receive positive reinforcement.

Let $x_i$ denote the generator input (here, $x_i \equiv T$ plus any formatting/context) and $y_i$ denote the generator output representing the auxiliary pair $(p_i,a_i)$. We compute GRPO-style advantages $\hat{A}_i$ computed from the rewards and normalized across the $m$ samples. We then optimize a clipped policy-gradient surrogate:
\[
\mathcal{L}_{\mathrm{aux}}(\theta)
=
-\mathbb{E}_{\{y_i\}\sim \pi_{\theta_{\mathrm{old}}}(\cdot\mid x)}
\left[
\frac{1}{m}\sum_{i=1}^{m}
\min\!\Big(
\frac{\pi_\theta(y_i\mid x_i)}{\pi_{\theta_{\mathrm{old}}}(y_i\mid x_i)}\,\hat{A}_i,\;
\mathrm{clip}(\frac{\pi_\theta(y_i\mid x_i)}{\pi_{\theta_{\mathrm{old}}}(y_i\mid x_i)},1-\epsilon,1+\epsilon)\,\hat{A}_i
\Big)
\right].
\]
To keep solving in-distribution, we include an additional term that reinforces solving performance to yield $\mathcal{L}_{\mathrm{generator}}(\theta) = \mathcal{L}_{\mathrm{aux}}(\theta) + \gamma\,\mathcal{L}_{\mathrm{solve}}(\theta),$
with hyperparameter $\gamma$. When gold solutions are available, $\mathcal{L}_{\mathrm{solve}}$ is an SFT cross-entropy loss on the gold solution. When only verification is available, $\mathcal{L}_{\mathrm{solve}}$ is a similar GRPO-style loss over the $k$ attempts on $T$ using binary verifier outcomes as rewards (and $\pi_{\theta_{\mathrm{old}}} := \pi_{\theta'}$ for that solve-policy update).

Computing $\nabla_\eta \mathcal{L}_{\mathrm{outer}}$ requires differentiating through the inner update(s), which na\"ively involves expensive second-order terms and large activation storage when backpropagating through an unrolled inner loop. We build on recent work on scalable bilevel differentiation \citet{MixFlow, DataRater}, computing meta-gradients in a memory-efficient mixed-mode form (forward-over-reverse vs. standard reverse-over-reverse unroll), as well as block-level rematerialization with gradient checkpointing. 

Each meta-training iteration then proceeds as: (1) generate data for $T$, (2) compute scores via $\eta$, (3) adapt model parameters on weighted data, (4) compute outer loss on $T$, (5) update both $\eta$ and $\theta$ using the meta-gradients.

\section{Experiments}
\subsection{Experimental Setup}
We work with the task of mathematical reasoning, evaluating performance on the MATH-500 benchmark \citep{MATH-500}, which contains mathematical reasoning problems from various different fields of math, measuring our methods' ability to adapt for different domains. We train on 1,000 examples sampled from the training split of the MATH dataset \citep{MATH}, disjoint from the evaluation tasks.  We note that we minimally optimize for actual math generation, and that our approach primarily leverages training data as signal to \textit{meta-learn}.  

We work with Llama 3.1-8B-Instruct \citep{grattafiori2024llama3herdmodels} as our base model and utilize parameter-efficient LoRA training \citep{LoRA} for temporary inner loop updates. The scorer model follows the base model's architecture but is lightweight in size and has a sigmoid-constrained numerical scoring head. We train for about 100 steps with 20 warmup steps where the generator is frozen and the scorer updated. We perform 2 inner steps and unroll through both. 
 
For each task, we generate 12 candidate synthetic training examples. For the verifier-only setting, the outer loss is computed by sampling 6 solution attempts on the target task. At test time we generate 6 synthetic training examples then perform an unweighted SFT LoRA update before attempting the target theorem. 

\subsection{Results}
We compare multiple baselines. Our results are shown in Table ~\ref{tab:results}.
\begin{itemize}
    \item \textbf{Base: } The base model (Llama-3.1-8B-Instruct).
    \item \textbf{Base Test-Time Self-Synthesis (TT-SS): } The base model generates $k$ synthetic training examples, performs a LoRA update, then attempts the target task.
    \item \textbf{Base Test-Time Training (TTT): } $k$ examples are sampled from the MATH training data; the base model performs a LoRA update, then attempts the target task.
    \item \textbf{Solver GRPO: } The base model trained with vanilla GRPO directly for math solving with binary correctness rewards.
    \item \textbf{\ours{}: } \ours{} generates $k$ synthetic training examples, performs a LoRA update, then attempts the target task (this represents the setting with no gold solution, only verification).
    \item \textbf{\ours{}$_{gold}$: } \ours{} generates $k$ synthetic training examples, performs a LoRA update, then attempts the target task (this represents the setting using the gold solution in the outer loss).
\end{itemize}

\begin{table}[h] 
    \centering 
    \caption{Accuracy of methods on MATH-500.} 
    \begin{tabular}{@{}lc@{}} 
    \toprule 
    \textbf{Method} & \textbf{MATH-500} \\ 
    \midrule Base & 43.6\% \\ 
    Base TTT & 41.2\% \\ 
    Base TT-SS & 46.6\% \\ 
    \midrule 
    Solver GRPO & 49.1\% \\ 
    \midrule 
    \textbf{\ours{}} & \textbf{59.0\%} \\ 
    \textbf{\ours{}$_{gold}$} \hspace{45pt} & 54.1\% \\ 
    \bottomrule 
    \end{tabular} 
\label{tab:results} 
\end{table}

In the evaluations in Table \ref{tab:results}, \ours{} achieves the strongest performance, substantially outperforming all baselines and improving on the Base model by 15.4pp (x1.35). Without meta-learning, the model shows limited capabilities in generating targeted beneficial synthetic data for self-adaptation (Base-TT-SS), but is able to improve from Base by 3.0pp. Naive TTT slightly hurts performance, suggesting that generic test-time updates without problem-specific supervision can introduce drift. \ours{} performs strongly in both settings where it updates according to golden solutions and verified self-generated solutions.  

\begin{figure}[h]
    \centering
    \includegraphics[width=0.6\linewidth]{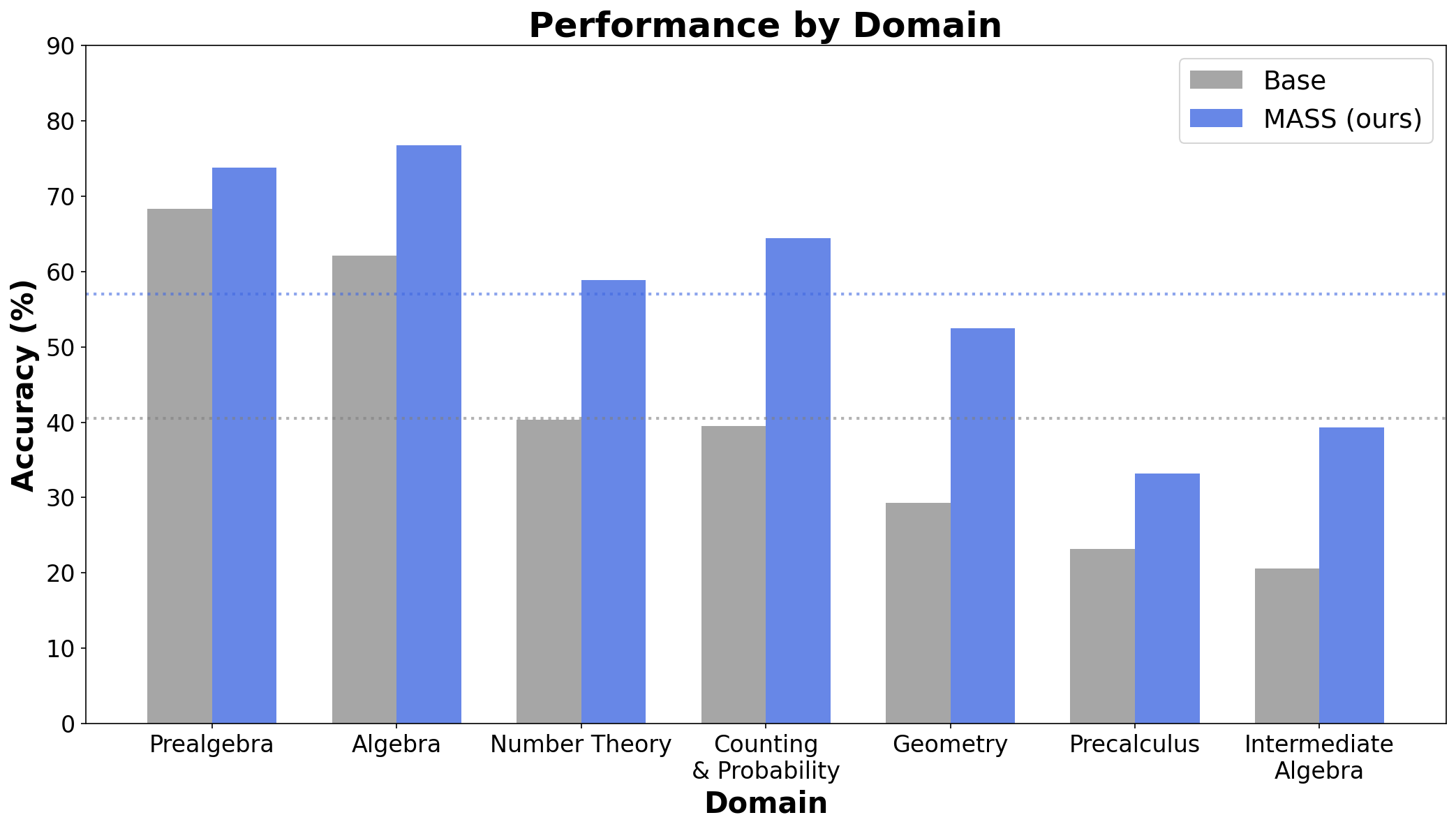}
    \caption{Performance gains by domain.}
    \label{fig:bars}
\end{figure}

\ours{} especially shows its potential in improving performance across domains. In Figure \ref{fig:bars}, it provides the greatest gains where initial performance is weakest (Intermediate Algebra, 1.92x) and overall improves consistency of performance across domains. \ours{} demonstrates that it can effectively utilize test-time compute to generate problem-specific synthetic data to adapt itself to the domain in a data-efficient manner, better leveraging its knowledge and resources for its environment.

\section{Conclusion}
Test-time \ours{} (\textsc{Meta-Adaptation with Self-Synthesis}) demonstrates that models can meta-learn to self-adapt at inference time by generating problem-specific synthetic curricula. \ours{} learns \emph{what} self-generated updates are useful for optimal performance gains on the current input, enabling generalizable per-instance self-improvement. We consistently improve over strong and compute-matched baselines in mathematical reasoning. Our work emphasizes synthetic data generation for self-improvement and efficient meta-learning for data attribution, and opens up the method as a general mechanism for models to robustly adapt in any setting.

\bibliography{references}
\bibliographystyle{iclr2026_conference}

\end{document}

%% file: math_commands.tex
\usepackage{amsmath,amsfonts,bm}

\def\eqref#1{equation~\ref{#1}}

\def\1{\bm{1}}

\DeclareMathAlphabet{\mathsfit}{\encodingdefault}{\sfdefault}{m}{sl}
\SetMathAlphabet{\mathsfit}{bold}{\encodingdefault}{\sfdefault}{bx}{n}



%% file: references.bib
@article{MATH,
  title={Measuring Mathematical Problem Solving With the MATH Dataset},
  author={Dan Hendrycks and Collin Burns and Saurav Kadavath and Akul Arora and Steven Basart and Eric Tang and Dawn Song and Jacob Steinhardt},
  journal={NeurIPS},
  year={2021}
}

@article{MATH-500,
  title={Let's Verify Step by Step},
  author={Lightman, Hunter and Kosaraju, Vineet and Burda, Yura and Edwards, Harri and Baker, Bowen and Lee, Teddy and Leike, Jan and Schulman, John and Sutskever, Ilya and Cobbe, Karl},
  journal={arXiv preprint arXiv:2305.20050},
  year={2023}
}

@misc{grattafiori2024llama3herdmodels,
      title={The Llama 3 Herd of Models}, 
      author={Aaron Grattafiori and Abhimanyu Dubey and Abhinav Jauhri and Abhinav Pandey et al.},
      year={2024},
      eprint={2407.21783},
      archivePrefix={arXiv},
      primaryClass={cs.AI},
      url={https://arxiv.org/abs/2407.21783}, 
}

@article{STaR,
  title        = {{ST}aR: Bootstrapping Reasoning With Reasoning},
  author       = {Zelikman, Eric and Wu, Yuhuai and Mu, Jesse and Goodman, Noah D.},
  journal      = {arXiv preprint arXiv:2203.14465},
  year         = {2022},
  url          = {https://arxiv.org/abs/2203.14465}
}

@inproceedings{SelfInstruct,
  title        = {{SELF-INSTRUCT}: Aligning Language Models with Self-Generated Instructions},
  author       = {Wang, Yizhong and Kordi, Yeganeh and Mishra, Swaroop and Liu, Alisa
                  and Smith, Noah A. and Khashabi, Daniel and Hajishirzi, Hannaneh},
  booktitle    = {Proceedings of the 61st Annual Meeting of the Association for Computational Linguistics (ACL)},
  year         = {2023},
  pages        = {13484--13508},
  url          = {https://aclanthology.org/2023.acl-long.754/}
}

@article{MAML,
  author       = {Chelsea Finn and
                  Pieter Abbeel and
                  Sergey Levine},
  title        = {Model-Agnostic Meta-Learning for Fast Adaptation of Deep Networks},
  journal      = {CoRR},
  volume       = {abs/1703.03400},
  year         = {2017},
  url          = {http://arxiv.org/abs/1703.03400},
  eprinttype    = {arXiv},
  eprint       = {1703.03400},
  bibsource    = {dblp computer science bibliography, https://dblp.org}
}

@article{STP,
  title   = {STP: Self-play LLM Theorem Provers with Iterative Conjecturing and Proving},
  author  = {Kefan Dong and Tengyu Ma},
  journal = {arXiv preprint arXiv:2502.00212},
  year    = {2025},
  url     = {https://arxiv.org/abs/2502.00212}
}

@article{SEAL,
  title   = {Self-Adapting Language Models},
  author  = {Adam Zweiger and Jyothish Pari and Han Guo and Ekin Aky{\"u}rek and Yoon Kim and Pulkit Agrawal},
  journal = {arXiv preprint arXiv:2506.10943},
  year    = {2025},
  url     = {https://arxiv.org/abs/2506.10943}
}

@article{MixFlow,
  title   = {Scalable Meta-Learning via Mixed-Mode Differentiation},
  author  = {Iurii Kemaev and Dan A. Calian and Luisa M. Zintgraf and Gregory Farquhar and Hado van Hasselt},
  journal = {arXiv preprint arXiv:2505.00793},
  year    = {2025},
  url     = {https://arxiv.org/abs/2505.00793}
}

@misc{DataRater,
      title={DataRater: Meta-Learned Dataset Curation}, 
      author={Dan A. Calian and Gregory Farquhar and Iurii Kemaev and Luisa M. Zintgraf and Matteo Hessel and Jeremy Shar and Junhyuk Oh and András György and Tom Schaul and Jeffrey Dean and Hado van Hasselt and David Silver},
      year={2025},
      eprint={2505.17895},
      archivePrefix={arXiv},
      primaryClass={stat.ML},
      url={https://arxiv.org/abs/2505.17895}, 
}

@InProceedings{TTT,
  title = 	 {Test-Time Training with Self-Supervision for Generalization under Distribution Shifts},
  author =       {Sun, Yu and Wang, Xiaolong and Liu, Zhuang and Miller, John and Efros, Alexei and Hardt, Moritz},
  booktitle = 	 {Proceedings of the 37th International Conference on Machine Learning},
  pages = 	 {9229--9248},
  year = 	 {2020},
  editor = 	 {III, Hal Daumé and Singh, Aarti},
  volume = 	 {119},
  series = 	 {Proceedings of Machine Learning Research},
  month = 	 {13--18 Jul},
  publisher =    {PMLR},
  url = 	 {https://proceedings.mlr.press/v119/sun20b.html}
}

@misc{LoRA,
      title={LoRA: Low-Rank Adaptation of Large Language Models}, 
      author={Edward J. Hu and Yelong Shen and Phillip Wallis and Zeyuan Allen-Zhu and Yuanzhi Li and Shean Wang and Lu Wang and Weizhu Chen},
      year={2021},
      eprint={2106.09685},
      archivePrefix={arXiv},
      primaryClass={cs.CL},
      url={https://arxiv.org/abs/2106.09685}, 
}

@misc{hu2025test,
      title={Test-Time Learning for Large Language Models}, 
      author={Jinwu Hu and Zhitian Zhang and Guohao Chen and Xutao Wen and Chao Shuai and Wei Luo and Bin Xiao and Yuanqing Li and Mingkui Tan},
      year={2025},
      eprint={2505.20633},
      archivePrefix={arXiv},
      primaryClass={cs.CL},
      url={https://arxiv.org/abs/2505.20633}, 
}

@article{zheng2024towards,
  title={Towards Lifelong Learning of Large Language Models: A Survey},
  author={Zheng, Junhao and Zheng, Chengming and Xue, Qianli and Guo, Yujun and Shi, Siyi and Zhou, Kaiwen and Xu, Xiao and He, Tao},
  journal={arXiv preprint arXiv:2406.06391},
  year={2024}
}

@article{wu2025continual,
  title={Continual Learning of Large Language Models: A Comprehensive Survey},
  author={Wu, Tongtong and Luo, Linhao and Li, Yuan-Fang and Pan, Shirui and Vu, Thuy-Trang and Haffari, Gholamreza},
  journal={ACM Computing Surveys},
  year={2025}
}
